\documentclass[runningheads,a4paper]{art}
\usepackage{makeidx}
\usepackage{graphicx}
\usepackage{xspace}
\usepackage[table]{xcolor}
\usepackage{tikz}
\usepackage[linesnumbered,ruled,vlined]{algorithm2e}
\usepackage{amsmath}
\usepackage{amssymb}
\usepackage{multirow}
\usepackage{array}
\usepackage{colortbl}
\usepackage{ctable}
\usepackage{epstopdf}
\usepackage{enumitem}
\usepackage{textcomp}
\usepackage{mathrsfs}
\usepackage[hidelinks]{hyperref}
\usepackage{xcolor}
\hypersetup{
	colorlinks,
	linkcolor={blue!80!black},
	citecolor={blue!80!black},
	urlcolor={blue!80!black}
}

\newcommand{\R}{\textup}
\newcommand{\GC}{\cellcolor{gray!10}}

\SetKwRepeat{Do}{do}{while}
\SetKwFunction{bestSplit}{bestSplit}
\SetKwFunction{merge}{merge}
\SetKwFunction{addAll}{addAll}
\SetKwFunction{mergePair}{mergePair}
\SetKwFunction{assoc}{assoc}
\SetKwFunction{corr}{corr}

\begin{document}

\title{Decision Stream:\protect\\ Cultivating Deep Decision Trees}

\author{Dmitry Ignatov\inst{1} \and Andrey Ignatov\inst{2}}

\authorrunning{D.Yu. Ignatov and A.D. Ignatov}
\tocauthor{Dmitry~Ignatov and Andrey~Ignatov}

\institute{Russian Research Center of Huawei Technologies, Russia\\
	\and
	ETH Zurich, Switzerland\\
	\ 
	\email{ignatov.dmitry@huawei.com, andrey.ignatoff@gmail.com}
}

\maketitle

\begin{abstract}
Various modifications of decision trees have been extensively used during the past years due to their high efficiency and interpretability. Tree node splitting based on relevant feature selection is a key step of decision tree learning, at the same time being their major shortcoming: the recursive nodes partitioning leads to geometric reduction of data quantity in the leaf nodes, which causes an excessive model complexity and data overfitting. In this paper, we present a novel architecture~--- a Decision Stream,~--- aimed to overcome this problem. Instead of building a tree structure during the learning process, we propose merging nodes from different branches based on their similarity that is estimated with two-sample test statistics, which leads to generation of a deep directed acyclic graph of decision rules that can consist of hundreds of levels. To evaluate the proposed solution, we test it on several common machine learning problems~--- credit scoring, twitter sentiment analysis, aircraft flight control, MNIST and CIFAR image classification, synthetic data classification and regression. Our experimental results reveal that the proposed approach significantly outperforms the standard decision tree learning methods on both regression and classification tasks, yielding a prediction error decrease up to 35\,\%.
\end{abstract}

\keywords{
decision tree, data fusion, two-sample test statistic, distributed machine learning.}	
\linebreak

\section{Introduction}

With the recent growth of data amount available for analysis and exploration, there is an inevitable need of comprehensive and automated methods for intellectual data processing. Decision tree (DT) is one of the most popular techniques in this area, and due to robustness and efficiency this prediction model became a standard tool for machine learning and big data problems. The idea behind this method is to separate one complex decision rule into a union of primitive rules, which leads to another crucial property~--- DT
can be easily interpreted by human compared to many other machine learning techniques.

The DT construction is performed by recursive data partitioning. At each stage the best splitting rule is determined, and data from the current node is divided into child nodes according to the selected criterion. The same procedure is recursively applied to all new nodes in the generated tree until the stopping condition is met. While being a fast and clear way of data splitting, the geometrical reduction of data quantity in the nodes leads to their exhaustion and causes poor generalization ability and data overfitting. Since multiple partitioning generates many nodes with the same or similar label distribution (especially in the lower layers), it looks quite natural to merge such nodes to diminish the problem of data exhaustion and to continually increase the purity of the separated samples.

\begin{figure}
	\centering
	\includegraphics[height=4.9cm]{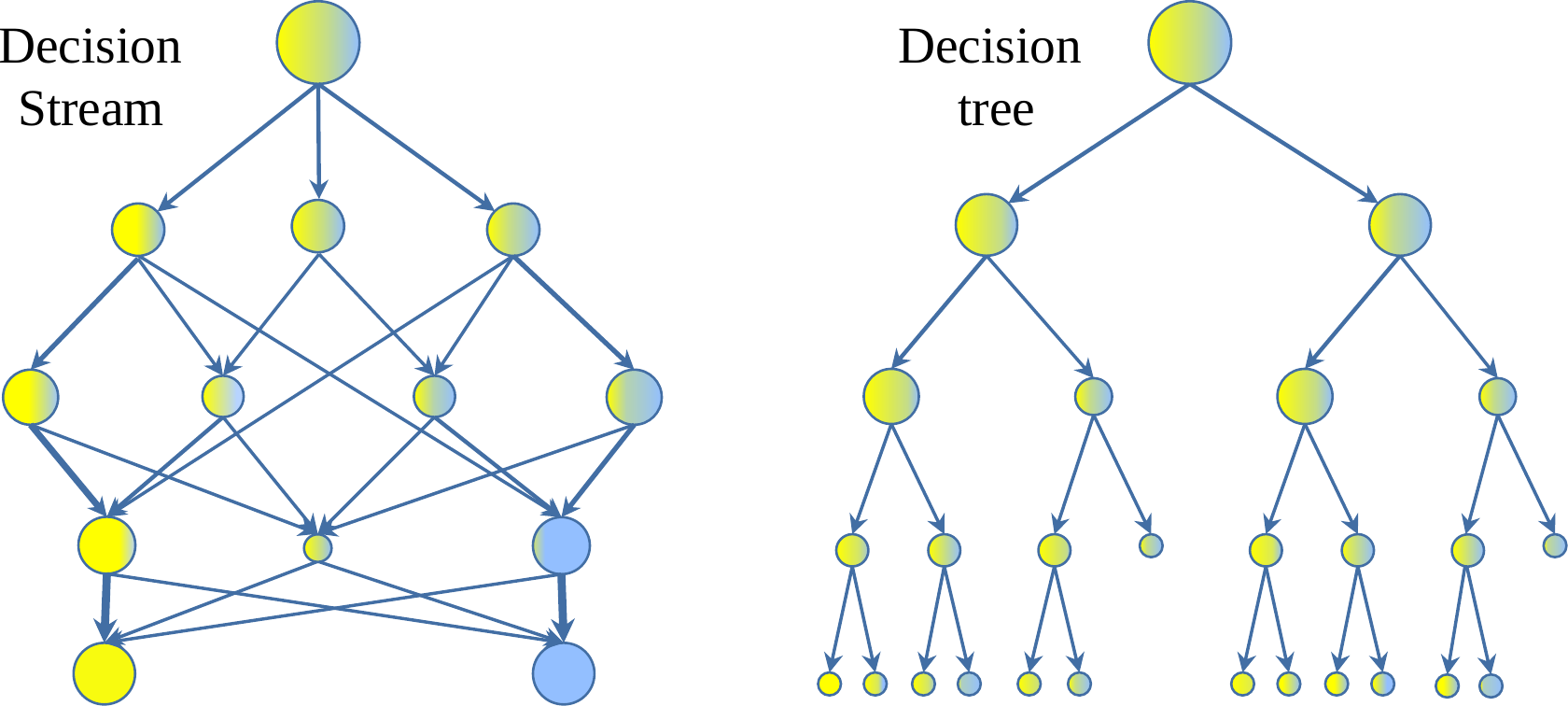}
	\caption[ ]{Predictive models: Decision Stream vs. decision tree.}\label{fig:model:comparison}
\end{figure}

In this paper, we propose a novel method for regression and classification tasks~--- a Decision Stream (DS), where decision branches are loosely split and merged like natural streams of a waterfall (Fig.~\ref{fig:model:comparison}). In contrast to the classical decision tree algorithm, the proposed method builds a deep directed acyclic graph with higher degree of connectivity by merging statistically indistinguishable nodes, which leads to reduction of the model width and better generalization due to more representative data samples. The split and merge operations are combined in this approach and repeated at each step of the iterative learning process. The performed experiments demonstrate that the proposed method achieves notably better results compared to the standard decision tree approach, at the same time showing high computational performance during training in distributed systems. The data and software related to this paper are available on GitHub\footnote{  \url{https://github.com/aiff22/Decision-Stream}}.

The rest of the paper is organized as follows. Section~\ref{relatedWork} gives an overview of the related works. Section~\ref{methodology} presents in details the proposed approach, and Section~\ref{Experiments} provides the experimental results obtained on the real-world problems as well as on synthetic data. Section~\ref{conclusion} summarizes our conclusions.

\section{Related Work}
\label{relatedWork}

Decision trees have been extensively studied, and a large number of their modifications were developed during the past years. The proposed methods include the Iterative Dichotomiser 3 and its successor~--- C4.5~\cite{Quinlan}, Classification and Regression Tree (CART)~\cite{Breiman:Friedman}, Chi-squared Automatic Interaction Detection (CHAID)~\cite{Kass}, Quick, Unbiased, Efficient, Statistical Tree (QUEST)~\cite{Loh} and various modifications of these algorithms~\cite{Loh}--\cite{Sanz}. Despite the essential difference in the training procedure, they usually tend to show similar performance on many real-world regression and classification tasks~\cite{Ture}--\cite{Pitombo}.

The majority of these algorithms consider only node partitioning for decision tree construction, or use node merging as an auxiliary procedure that has no significant effect on the tree structure. For instance, C4.5 and CART algorithms as well as their modifications~\cite{Loh}--\cite{Sanz} perform only node splitting based on the selected features without any merging or fusion operations. QUEST algorithm merges several classes into two superclasses to produce one binary split~\cite{Loh:Shih}. In~\cite{Ciampi}, the number of terminal nodes is reduced by fusing the leaves with similar predictions after the training is finished. The CHAID algorithm merges data samples within a node, which is equivalent to using a modified splitting criterion. Data samples are fused based on the significance of their similarity estimated by test statistics: $\chi^2$ test~\cite{Kass} for categorical label and F-test~\cite{Nisbet} for continuous.

A fundamentally different approach based on Occam's razor concept was proposed for decision tree size reduction in~\cite{Oliver}, where decision graph is constructed on the basis of hill climbing heuristic by merging nodes from adjacent levels according to minimum message length principle with goal to produce a model of minimum size while preserving/increasing its accuracy. This technique has demonstrated an advantage over standard decision trees in experiments~\cite{Kohavi}--\cite{Shotton}.

In this work, we present a Decision Stream algorithm that combines the classical decision tree learning method with a new procedure~--- statistically-based merging of nodes from the same and/or different levels of DS. The predictive model is growing till no improvements are achievable, considering different data recombinations, and resulting in deep directed acyclic graph architecture and statistically-significant data partition.
\newpage
\section{Decision Stream}
\label{methodology}
In this section, we describe the proposed Decision Stream algorithm. The main concept of this method consists in merging similar nodes after each splitting iteration. The similarity is estimated using two-sample test statistics that is applied to compare the distribution of labels in each pair of nodes. The nodes are merged if the difference is statistically insignificant. This procedure eliminates the classical problem of decision trees~--- progressive decrease of data quantity in the leaf nodes, and produces a more general structure~--- a directed acyclic graph (Fig.~\ref{fig:model:comparison}), which can be extremely deep. A more detailed explanation of the algorithm is provided below.

\begin{figure*}[t]
	\centering{
	\includegraphics[height=4.0cm]{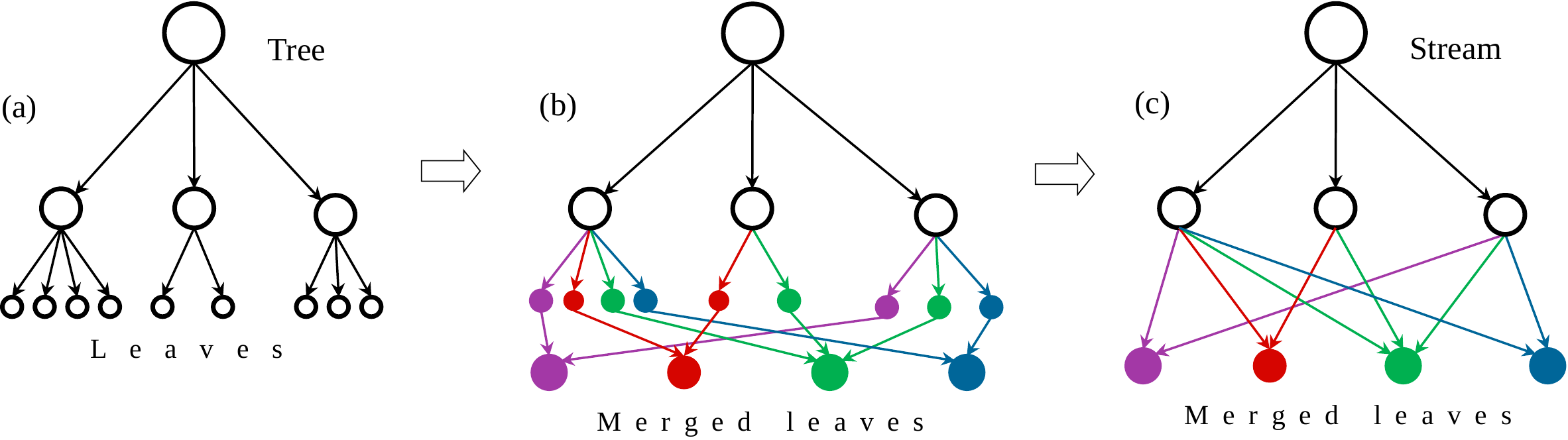}
	\caption{Merging leaf nodes in Decision Stream.}\label{fig:leaf:merging}}
\end{figure*}

\subsection{Node Merging with Two-Sample Test Statistics}

The overview of the merging operation is illustrated in the Fig.~\ref{fig:leaf:merging}. After the classical decision tree branching, the merging algorithm takes as an input leaf nodes generated at the current stage (Fig.~\ref{fig:leaf:merging}(a)) as well as previously obtained unsplit leaves from the upper levels of the model, and fuses statistically similar nodes (Fig.~\ref{fig:leaf:merging}(b-c)) using an input parameter~--- significance threshold $P_{\R{lim}}$. Since the nodes are merged based on the similarity of their label distributions, the merging procedure can be considered as the statistically-based label clustering.


Merging Algorithm~\ref{algo:leaf:merging} consists of an outer and inner loop. In the outer loop the leaves are sorted in ascending order according to the number of associated samples. The inner loop consists of the following three steps:

\begin{algorithm}[h]
	\caption{Merging the Leaf Nodes}\label{algo:leaf:merging}
	\DontPrintSemicolon
	\KwData{A set $A=\{node_0, node_1, \ldots, node_n\}$ of leaves}
	\KwIn{$P_{\R{lim}}$}
	\KwResult{A set of merged leaves}
	\SetKwFunction{remove}{remove}
	\SetKwFunction{poll}{poll}
	\SetKwFunction{add}{add}
	
	\Do{$\R{size of } A \R{ is decreased}$}{
		$A' \gets \emptyset$
		
		sort nodes of $A$ by $\#$ samples in ascending order
		
		\Do{$A \ne \emptyset$}{
			$node_t \gets A.\poll()$
			
			$(node_t, node_x) \gets \mathop{\arg\max }\limits_{node_i \in A} S_{\R{st}}(node_t, node_i)$
			
			$p \gets S_{\R{st}}(node_t, node_x)$
			
			\uIf{$p > P_{\R{lim}}$}{
				
				$A.\remove(node_x)$
				
				$A'.\add(\mergePair(node_t, node_x))$
			}
			\Else{
				$A'.\add(node_t)$
			}
		}
		$A \gets A'$
	}
	\Return{$A$}
\end{algorithm}

\begin{enumerate}
 	\item Leaf $node_t$ is picked up from the head of the sorted collection.
 	\item For each ($node_t$, $node_i$) pair we compute the similarity of two nodes and then take the leaf $node_x$ that corresponds to its highest value. The similarity is calculated by the function $S_{\R{st}}(node_i, node_j)$ with two-sample test statistics~(\ref{sec:splitting:merging:criteria}). Function  $S_{\R{st}}$ returns the significance level $p$ representing the probability that the mean values of labels associated with these two nodes are identical.
\item If the obtained significance level $p$ is above the threshold $P_{\R{lim}}$, the leaves $node_t$ and $node_x$ are merged into a new leaf with parents obtained by uniting the parents of the merged nodes.
\end{enumerate}

\subsection{Decision Stream Training}

The whole DS training procedure is described in Algorithm~\ref{algo:decision:stream}, where each learning iteration consists of two steps. At the first step, DS grows using the classical decision tree branching operation~--- new nodes are created by splitting all current non-terminal leaves~\cite{Breiman:Friedman,Loh,Glavan}. At the second step, the leaves are merged using the procedure described in Algorithm~\ref{algo:leaf:merging}. A leaf is marked as terminal if it cannot be split into statistically different child nodes. The pair of splitting and merging steps is iteratively performed till the stopping criterion is met. If all leaves are terminal or the prediction accuracy is not improved, the DS training is finished and Algorithm~\ref{algo:decision:stream} returns the reference to the root node of the generated DS. To estimate the prediction accuracy, we use a cross-node Gini impurity measure calculated for $K$ leaf nodes and $J$ classes:
\begin{equation}
I_{\R{G}}(nodes) = \sum^K_{k = 1}{\frac{n_k}{N} \sum^J_{j = 1}{f_{jk}(1 - f_{jk})}}\ ,
\end{equation}
where $N$ and $n_k$ is the number of samples in all leaves and leaf node $k$, respectively; $f_{jk}$ is the fraction of samples of class $j$ in leaf $k$.
\begin{algorithm}
	\caption{Decision Stream Training}\label{algo:decision:stream}
	\DontPrintSemicolon
	\KwData{A set $G_{\R{in}} = \{s_0, s_1, \ldots, s_n\}$ of labeled samples}
	\KwIn{Significance threshold $P_{\R{lim}}$}
	\KwResult{Decision Stream}
	
	\SetKwFunction{node}{node}
	\SetKwFunction{add}{add}
	
	$rootNode \gets \node(G_{\R{in}})$
	
	$A \gets \{rootNode\}$
	
	\Do{$ \exists\ \R{non-terminal } node \in A \R{ AND } I_{\R{G}}(A) \R{ is decreased}$}{
		
		$A' \gets \emptyset$
		
		\ForEach{$node \in A$}{
			\uIf{$node \R{ is terminal}$}{$A'.\add(node)$}
			\Else{
				$children \gets \bestSplit(node,  P_{\R{lim}})~\hspace{4.57cm}\triangleleft $~Algorithm~\ref{algo:compact:splitting}~or~\ref{algo:large:splitting}
				
				\uIf{$\#\ children \leq 1$}{
					
					mark $node$ as terminal
					
					$A'.\add(node)$
				}
				\Else{
					$A'.\addAll(children)$			
				}
		}}	
		
		$A \gets \merge(A', P_{\R{lim}})\hspace{7.73cm} \triangleleft $ Algorithm~\ref{algo:leaf:merging}
	}	
	\Return{$rootNode$}
\end{algorithm}
\newpage
\subsection{Splitting/Merging Criteria}
\label{sec:splitting:merging:criteria}

The splitting and merging operations are performed according to significance threshold $P_{\R{lim}}$. We take as the null hypothesis that labels of two nodes are from the same distribution and have the same mean value. The null hypothesis is rejected at the significance level $P_{\R{lim}}$, and in case of rejection we consider that the nodes are statistically different.
The similarity is estimated by function $S_{\R{st}}$ with pair of two-sample test statistics. We use Z-test/Student's t-test for labels with presumably normal  distribution. The choice between the tests is determined according to rule~\cite{Sprinthall}: Z-test is applied if the size of both data samples is greater than 30, Student's t-test~--- otherwise. For labels with non-normal distribution we use Kolmogorov-Smirnov/Mann-Whitney U tests: the first one is applied if the size of data samples is greater than 2, the second~--- otherwise. We prefer Kolmogorov-Smirnov over Mann-Whitney U test since it is more sensitive to variations of both location and shape of the empirical cumulative distribution function~\cite{Corder}, and provides better prediction accuracy in our experiments.

\bigskip

We propose two different versions of the split function $\bestSplit$: one for relatively small datasets, where a precise selection of the split is crucial; and one for large-scale datasets where a trade-off between the accuracy and running time is important due to big amount of training samples.

\subsection{Node Splitting for Non-Distributed Data}
\label{sec:compact:algorithm}

For non-distributed datasets the splitting is performed according to Algorithm~\ref{algo:compact:splitting}, which takes as an input the significance threshold $P_{\R{lim}}$ and a particular $node$. Firstly, binary splits of the data within the $node$ is generated for each unique value of every feature. Then the similarity function $S_{\R{st}}$ is calculated for each split, and the one with the lowest significance of similarity is selected. If this significance is smaller than the input threshold $P_{\R{lim}}$, the selected best split is returned, otherwise~--- splitting is rejected and the node becomes terminal. Though this method is rather computationally expensive, it provides the best split quality and is reasonable for compact datasets.

\begin{algorithm}
	\caption{Splitting the Non-Distributed Data}\label{algo:compact:splitting}
	\DontPrintSemicolon
	\KwData{A $node$ of Decision Stream}
	\KwIn{Significance threshold $P_{\R{lim}}$}
	\KwResult{Child nodes}
	
	$splits \gets$ all binary splits of data for all features within the $node$

	$(child_0, child_1) \gets \mathop{\arg\min }\limits_{(child_0^j, child_1^j) \in splits}S_{\R{st}}(child_0^j, child_1^j)$

	$p \gets S_{\R{st}}(child_0, child_1)$

	\uIf{$p < P_{\R{lim}}$}{
		\Return{$\{child_0, child_1\}$}
	}
	\Else{

		\Return{$\emptyset$}
	}
\end{algorithm}

\subsection{Node Splitting for Distributed Data}
\label{sec:large:algorithm}

\begin{figure*}
	\centering
	\includegraphics[height=3.5cm]{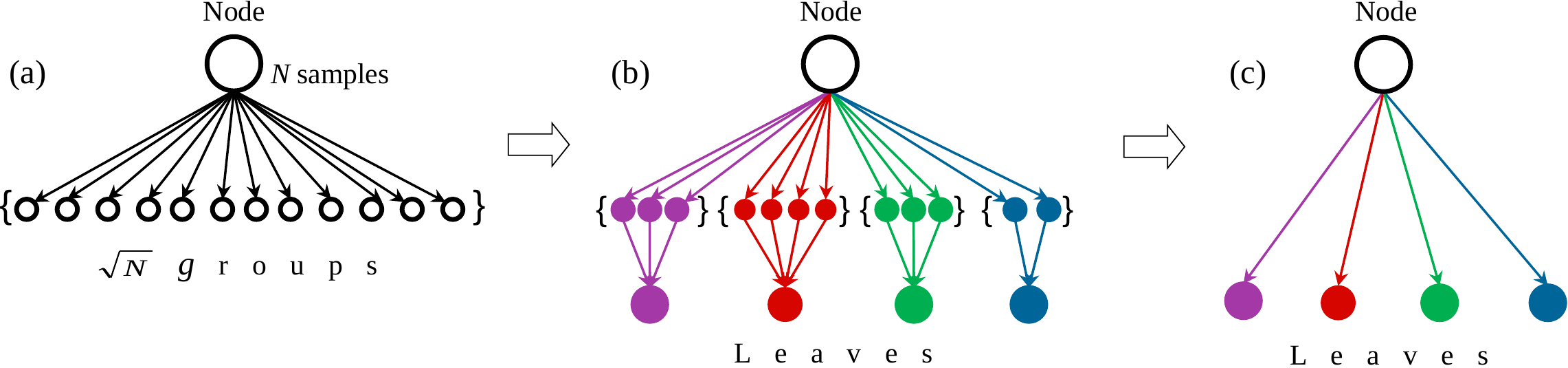}
	\caption[ ]{Data splitting (a) and fusion (b-c) within node of Decision Stream.}\label{fig:multiple:splitting}
\end{figure*}

Using the above algorithm for large-scale datasets is infeasible in most cases, thus we propose a different way of split selection designed for big data solutions. Instead of the greedy search, we perform data splitting based on the feature that is most correlated with label within a particular node~\cite{Salehi}. Another difference of the proposed method is that it attempts to produce multiple leaves for each node as shown in Fig.~\ref{fig:multiple:splitting}, so far as the large number of samples presumes the robustness of such split.

Algorithm~\ref{algo:large:splitting} demonstrates the body of the method. The procedure starts with function $\corr$ that selects the feature that is most correlated with the label. The obtained feature is then used to split the samples in the current node. If the feature is categorical, the samples are split by its categories, each one forming a leaf node. If the feature is continuous, all samples are firstly sorted according to values of the feature and then divided into $\sqrt{N}$ ranges, where $N$ is a number of samples in the node. Samples from the same range are then associated with one leaf node (Fig.~\ref{fig:multiple:splitting}(a)). At the next step, the adjacent leaves are merged using Algorithm~\ref{algo:leaf:merging} with threshold $P_{\R{lim}}$ until all neighboring nodes are statistically distinguishable (\mbox{Fig.~\ref{fig:multiple:splitting}(b-c)}). Finally, as soon as splitting with regard to categorical or continuous feature is finished, the obtained leaf nodes are merged again (this time not only adjacent ones) and the leaves providing statistically different predictions are returned.

The strength of correlation between the feature and label is estimated by function $\corr$ as described in Algorithm~\ref{algo:association}: if the feature and label are continuous, the correlation strength is calculated as coefficient of determination:
\begin{equation}
 r^2_{xy} = \left(\frac{\sum^n_{i=1}{(x_i - \bar{x})(y_i - \bar{y})}}{\sqrt{\sum^n_{i=1}{(x_i - \bar{x})^2}}\sqrt{\sum^n_{i=1}{(y_i - \bar{y})^2}}}\right)^2,
\end{equation}
otherwise it is computed as correlation ratio:

\begin{equation}
 \eta{^2_{y\mid x}} = \frac{\sum_x{n_x(\bar{y_x} - \bar{y})^2}}{\sum_{x,i}{(y_{x,i} - \bar{y})^2}}\ .
\end{equation}

Since both coefficients measure the same characteristics in discrete and continuous cases, we can compare the values obtained for different types of features to select the best one.

\begin{algorithm}
	\caption{Splitting the Distributed Data}\label{algo:large:splitting}
	\DontPrintSemicolon
	\KwData{A $node$ of Decision Stream}
	\KwIn{Significance threshold $P_{\R{lim}}$}
	\KwResult{Child nodes}	
	$feature \gets \mathop{\arg\max }\limits_{ft_j \in features} \corr(ft_j, lable)$
	
	\uIf{$feature \R{ is categorical}$}{
		$leaves \gets$ split $node$ samples with regard to categories of $feature$
	}
	\Else{
		
		$n \gets \sqrt{\#\R{ samples in the }node}$
		
		$leaves \gets $split $node$ samples with regard to $n$ ranges of $feature$
		
		\Do{$\R{quantity of } leaves \R{ is decreased}$}{
			$leaves \gets \merge$ adjacent $ leaves$ with $P_{\R{lim}}\hspace{4.7cm} \triangleleft $ Algorithm~\ref{algo:leaf:merging}
		}
	}
	$leaves \gets \merge(leaves, P_{\R{lim}})\hspace{7.65cm} \triangleleft $ Algorithm~\ref{algo:leaf:merging}
	
	\Return{$leaves$}
\end{algorithm}

\begin{algorithm}[t]
	\caption{Estimation of Correlation Strength}\label{algo:association}
	\DontPrintSemicolon
	\KwData{Variables $x$ and $y$}
	\KwResult{Strength of the correlation between variables $x$ and $y$}
	
	\uIf{$x \R{ is categorical}$}{
		\Return{$\eta{^2_{y\mid x}}$}
	}
	\ElseIf{$y \R{ is categorical}$}{
		\Return{$\eta{^2_{x\mid y}}$}
	}
	\Else{
		\Return{$r^2_{x,y}$}
	}
\end{algorithm}
\newpage
\section{Experiments}
\label{Experiments}

In this section, we describe the experiments conducted to evaluate the performance of the proposed Decision Stream algorithm. The solution was tested on five common machine learning problems, and on large-scale synthetic classification/regression data.

\subsection{Datasets}
\label{Dataset}

\begin{itemize}[label=$\bullet$]
	
	\item \textit{Credit scoring}\footnote{\url{https://www.kaggle.com/c/GiveMeSomeCredit/data/}}~--- \small{classification problem, 2 classes, 10 features, 100K training and 20K test samples.}\normalsize
	
	\smallskip
	
	\item \textit{Twitter sentiment analysis}\footnote{\url{http://alt.qcri.org/semeval2015/task10/}}~--- \small{classification problem, 3 classes (positive, negative, neutral), 500 features, 6500 training and 824 test samples. Features were generated using the bag-of-words model.}\normalsize
	
	\smallskip
	
	\item \textit{F16 aircraft control problem (Ailerons)}\footnote{\url{http://www.dcc.fc.up.pt/\textasciitilde ltorgo/Regression/DataSets.html}}~--- \small{regression problem, 40 features, 7154 training and 6596 test samples.}\normalsize
	
	\smallskip
	
	\item \textit{MNIST handwritten digits classification}\footnote{\url{http://yann.lecun.com/exdb/mnist/}}~--- \small{10 classes, 784 features, 60K training and 10K test samples.}\normalsize
	
	\smallskip
	
	\item \textit{CIFAR-10 image classification}\footnote{\url{https://www.cs.toronto.edu/\textasciitilde kriz/cifar.html}}~--- \small{10 classes, 1024 features, 50K training and 10K test samples. Features were extracted from the last convolutional layer of the pre-trained ResNet-18~\cite{He} CNN.}\normalsize
\end{itemize}

To tune model parameters, the training data for each problem was split into training (90\,\%) and validation (10\,\%) subsets. The same data was used for training and testing both decision tree and Decision Stream algorithms.

To get the baseline accuracy, we used the \textit{Scikit-learn}\footnote{\url{http://scikit-learn.org} (v. 0.18.1)} implementation of decision trees that provides four splitting criteria (information gain, Gini impurity, variance reduction, mean absolute error) and supports pre-pruning procedure. The parameters of DT were tuned for each dataset, including best criterion selection and tree pre-pruning.

Additionally, DS and DT algorithms were tested on large-scale synthetic classification and regression data generated on the fly by Spark Performance Tests Suite\footnote{\url{https://github.com/databricks/spark-perf/} (v. 1.6)}. Each generated sample consisted of 500 features (125~binary, 125~categorical with 20~categories and 250~continuous within interval [0, 1]) and represented binary classification and regression problems. The detailed classification and regression results are provided below.

\subsection{Tuning the Significance Threshold}

Significance threshold is the key parameter of DS algorithm, and in the first experiment our goal was to estimate its optimal value for each problem. The level of $P_{\R{lim}}$ was tuned as follows: for each dataset we varied it between $10^{-4}$ and 0.5 and for each value estimated the accuracy of DS on the validation set. For synthetic data the similarity of labels was estimated by unpaired two-sample Z-test and Students t-test, for all other datasets~--- by Kolmogorov-Smirnov and Mann-Whitney U nonparametric tests. For classification problems we use the standard accuracy metric, for regression tasks~--- the weighted absolute percentage error:

\begin{equation}
\label{equ:mean:percent}
\mathscr{L}_{M}(\textbf{X}, \textbf{Y}) = \frac{100}{\left|\sum^N_{i = 1}{y_i}\right|}\sum^N_{i = 1}{\left|y_i - \mathscr{F}(x_i)\right|}\ ,
\end{equation}
where \textbf{X} and \textbf{Y} are validation samples and their corresponding labels, $y_i$ and $\mathscr{F}(x_i)$ are the label and the prediction for sample $x_i$, respectively.

\begin{figure}
	\centering
	\includegraphics[height=4.9cm]{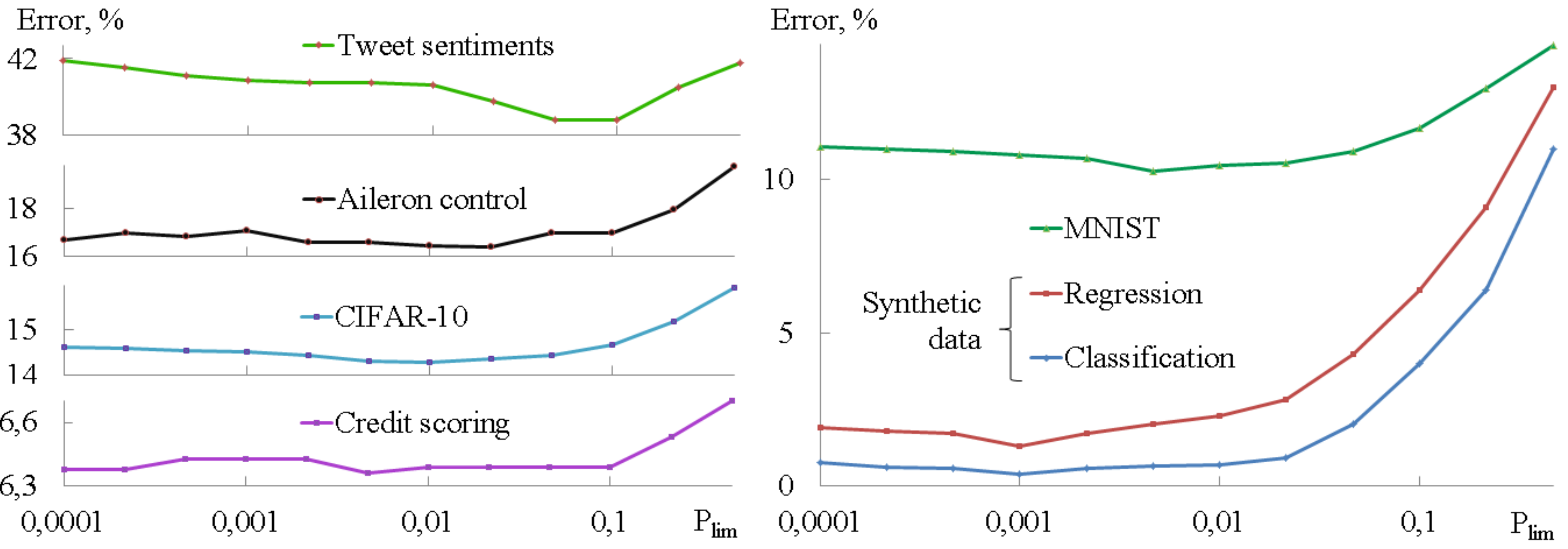}
	\caption[ ]{Dependency between the validation error and the value of significance threshold $P_{\R{lim}}$ on six datasets for Decision Stream algorithm.}\label{fig:significance:level}
\end{figure}

The results of the experiment are presented in Fig.~\ref{fig:significance:level}. The best accuracy was achieved at the significance threshold $P_{\R{lim}}$ that is equal to 0.005 for credit scoring, 0.05 for tweets, 0.02 for aileron control, 0.005 for MNIST, 0.01 for CIFAR-10 and 0.001 for synthetic data. The obtained values were used for DS training in the following experiments.

\subsection{Classification and Regression Results for Non-Distributed Data}

This section presents the results obtained using a Decision Stream implementation for non-distributed data. Along with the single DS and DT models, we train their ensembles generated using five methods: random forest~\cite{Ho}, extremely randomized trees~\cite{Geurts}, gradient boosting~\cite{Friedman} and bagging~\cite{Breiman}. Table~\ref{table:error:compacte:single} shows the results for single DS, DT and DS$^{-\R{merge}}$ models, where the last one denotes a DS with disabled merging phase. 
\begin{table}
	\caption{The prediction error (\%) of decision tree (DT), Decision Stream (DS) and DS with disabled merging phase (DS$^{-\R{merge}}$) on five benchmark datasets.}\label{table:error:compacte:single}
	\centering
	\newcolumntype{C}{ >{\centering\arraybackslash} m{2.4cm} }
	\newcolumntype{A}{ >{\centering\arraybackslash} m{2.7cm} }
	\begin{tabular}{m{1.5cm} C A C C A}\arrayrulecolor{gray}
		\hline
		Model & Credit scoring & Tweet sentiments
		& Aileron control
		& MNIST
		& CIFAR-10\\
		\specialrule{.1em}{0em}{0em}
		DT& 9.73 & 45.2 & 25.5 & 12.5 & 13.9 \\
		DS$^{-\R{merge}}$ & \textbf{6.33} & 39.9 & 18.2 & 25.0 & 19.7\\
		DS & 6.36 & \textbf{38.8} & \textbf{16.4} & \textbf{10.3} & \textbf{13.8} \\		
		\hline
	\end{tabular}
\end{table}

\begin{figure*}
	\centering
	\includegraphics[height=10.9cm]{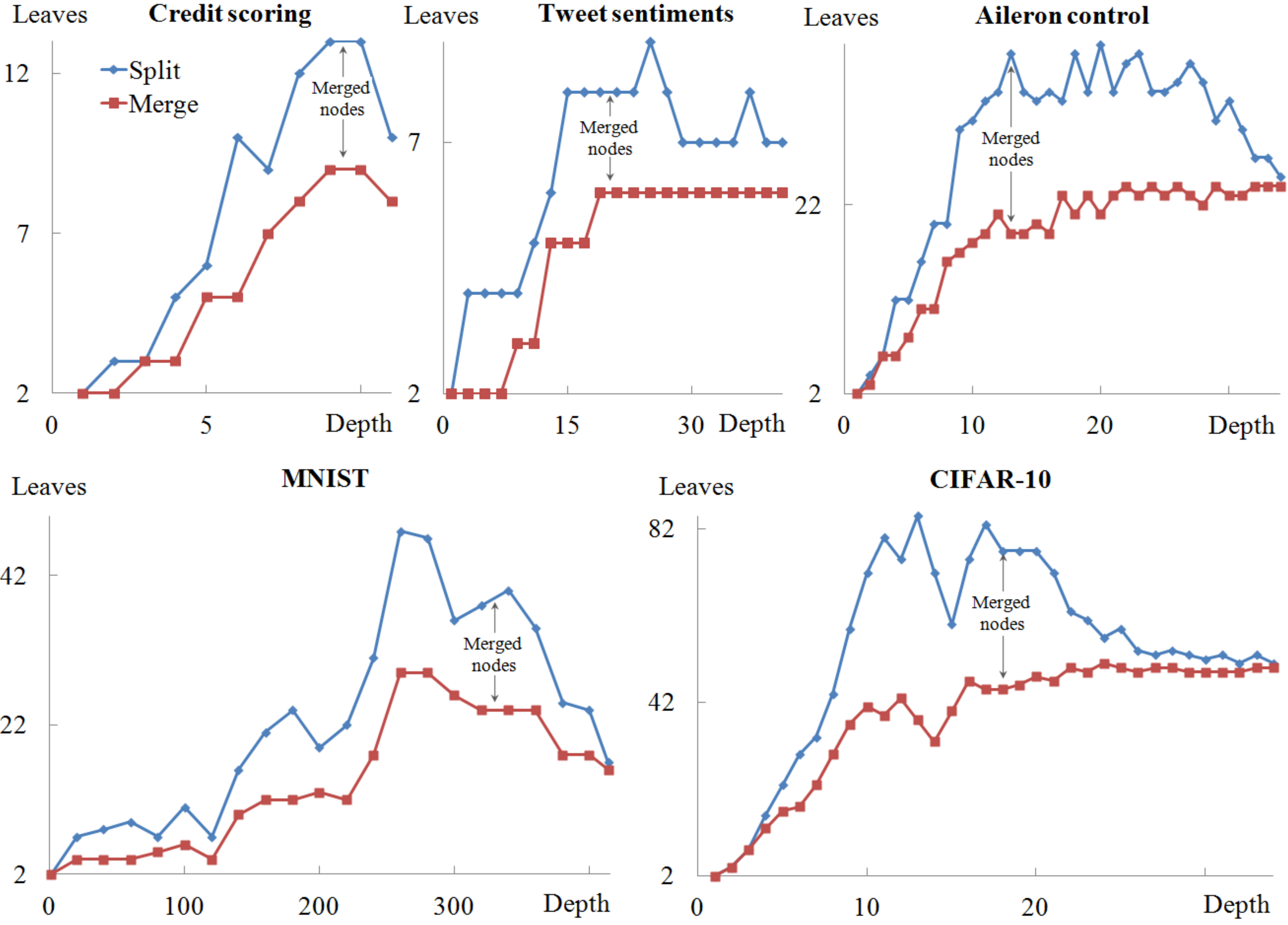}
	\caption[ ]{The number of leaf nodes after each splitting and merging operation over the growth of Decision Stream. Split operation always precedes the merge.}\label{fig:merging:dynamics}
\end{figure*}
We should note that DS$^{-\R{merge}}$ is not equivalent to DT since in this version node splitting is performed only if the resulting child nodes are statistically distinguishable. The results demonstrate that disabling of merging phase leads to substantially different accuracy~--- while on the first dataset with relatively low  complexity (2 classes, 10 features) it prevents minor overfitting, for other datasets with higher complexity (3--10 classes or continuous label, 40--1024 features) it results in an oversimplified tree model. Enabling the merging operation changes the situation: the growth doesn't stop on the stage of simple predictive model that has many similar leaf nodes~--- merging operation fuses them and thus forces the training procedure to continue that can result in very deep decision graphs. Fig.~\ref{fig:merging:dynamics} illustrates this oscillating behavior: the merge operation is performed till no more statistically distinguishable nodes can be produced. Table~\ref{table:error:compacte:single} demonstrates that this leads to significantly higher accuracy compared to the standard decision tree architecture: the error on the first four datasets is reduced by 34\,\%, 14\,\%, 35\,\% and 17\,\%, respectively.

\begin{figure}
	\centering
	\includegraphics[height=15.1cm]{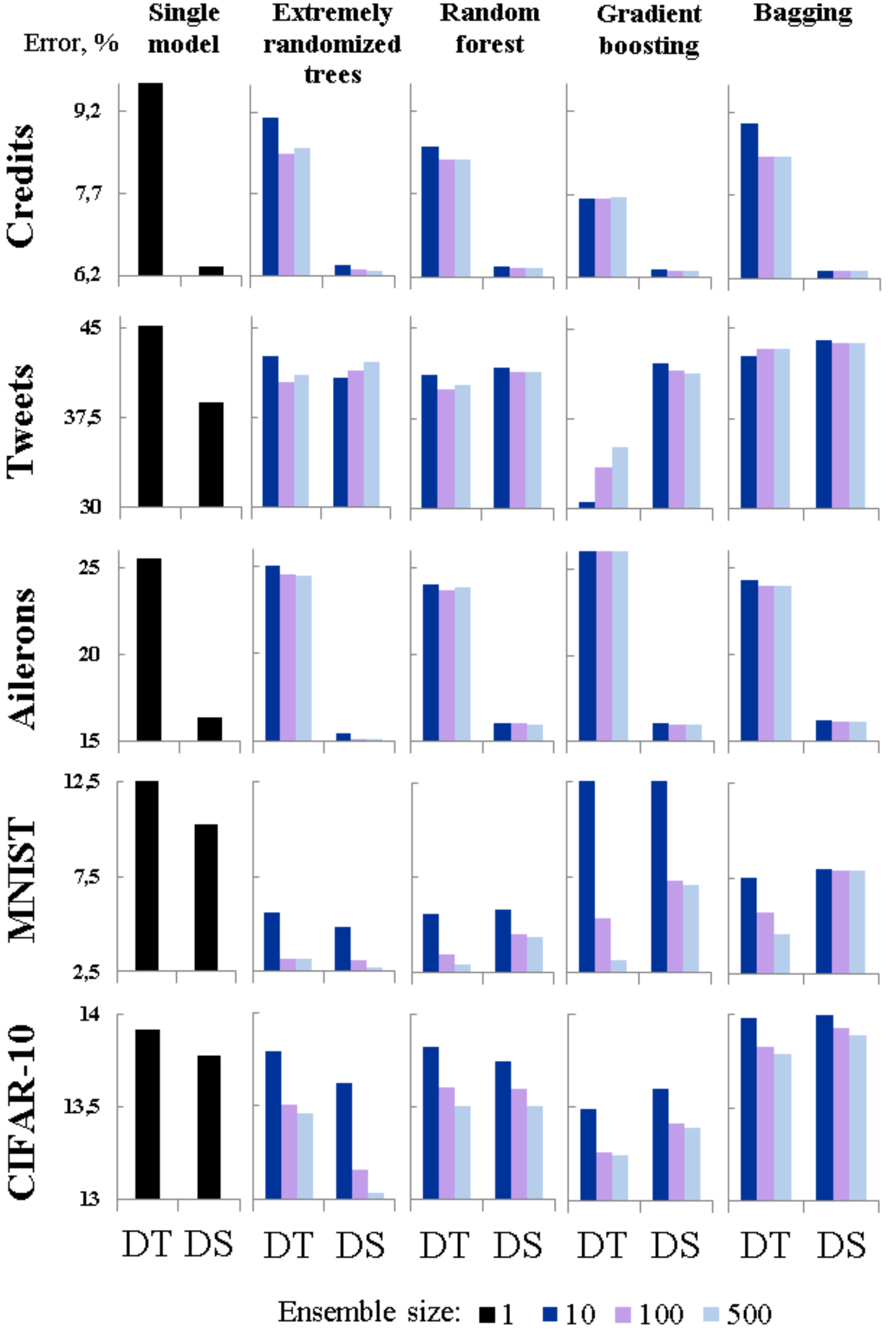}
	\caption[ ]{Prediction error (\%) for different sizes of ensembles constructed from decision trees (DT) and Decision Streams (DS).}\label{fig:error:mnist:cifar}
\end{figure}

Fig.~\ref{fig:error:mnist:cifar} illustrates the dependency between the size and the predictive error of ensembles constructed from decision trees and Decision Streams. The best results for all datasets are summarized in Table~\ref{table:error:compacte:ensembles}. 

\begin{table}
	\caption{The best results for decision tree (DT) and Decision Stream (DS) ensembles trained on five benchmark datasets.}\label{table:error:compacte:ensembles}
	\centering
	\newcolumntype{C}{ >{\centering\arraybackslash} m{1.70cm} }
	\newcolumntype{G}{ >{\centering\arraybackslash} m{2.85cm} }
	\newcolumntype{W}{ >{\centering\arraybackslash} m{2.5cm} }	
	\newcolumntype{A}{ >{\arraybackslash} m{1.0cm} }
	\begin{tabular}{A m{1.5cm} | G W C G C}\arrayrulecolor{gray}
		\hline
		Model & & Credit scoring & Tweet sentiments & CIFAR-10
		& Aileron control
		& MNIST	\\
		\specialrule{.1em}{0em}{0em}
		\multirow{2}{4em}{DT} & Method & \multicolumn{3}{c|}{Gradient boosting} &  \multicolumn{2}{c}{Random forest} \\
		& Error, \% & 7.62 & \textbf{30.4} &\multicolumn{1}{c|}{13.2} & 23.9 & 2.91 \\
		\hline
		\multirow{2}{4em}{DS} & Method & \multicolumn{5}{c}{ Extremely randomized trees }  \\
		& Error, \% & \textbf{6.31} & 38.8 & \textbf{13.0} & \textbf{15.0} & \textbf{2.66}\\
		\hline
	\end{tabular}	
\end{table}
As one can see, in all cases the best performance of Decision Stream ensemble was obtained when using the extremely randomized trees algorithm. The explanation of this effect is the following: in contrast to decision trees, the construction of Decision Streams involves a large number of recombinations caused by continuously repeating splitting and merging operations. The chances that DS will find the optimal solution are therefore higher compared to DT, but at the same time the resulting Decision Streams tend to provide less diverse results. The power of ensemble significantly depends on the diversity of predictors, which is thus lower in case of Decision Streams. Extremely randomized trees method partially solves this problem by using random features for training the DS, and therefore it tends to provide better final results compared to other methods.

In almost all cases the best results for Decision Stream are achieved by ensembles of size 500, with the only exception for  twitter sentiment analysis problem. The greatest advantage of DS over the DT is obtained on the credit scoring and aileron control tasks: a single DS outperforms all DT ensembles. Overall, the Decision Stream based methods have shown the best results on four out of five datasets with an average advantage of 16\,\%.

\newpage
\subsection{Classification and Regression Results for Large-Scale Data}

The next set of experiments is conducted using Apache Spark-based\footnote{\url{http://spark.apache.org} (v. 1.6)} distributed realization of Decision Stream and decision tree algorithms. For the last one an open-source implementation from MLlib machine learning library is used. To perform the distributed computations, the models were running on a computer cluster with 4 nodes (48 executors), 12 cores and 50 GB of RAM per node. The algorithms were trained on synthetic data generated by Spark Performance Tests Suite for classification and regression problems.

\newcommand{\tikzcircle}[2][gray,fill=white]{\tikz[baseline=-0.7ex]\draw[#1,radius=#2] (0,0) circle ;}
\definecolor{DSColor}{RGB}{255,101,101}
\begin{figure}[t]
	\centering
	\includegraphics[height=4.6cm]{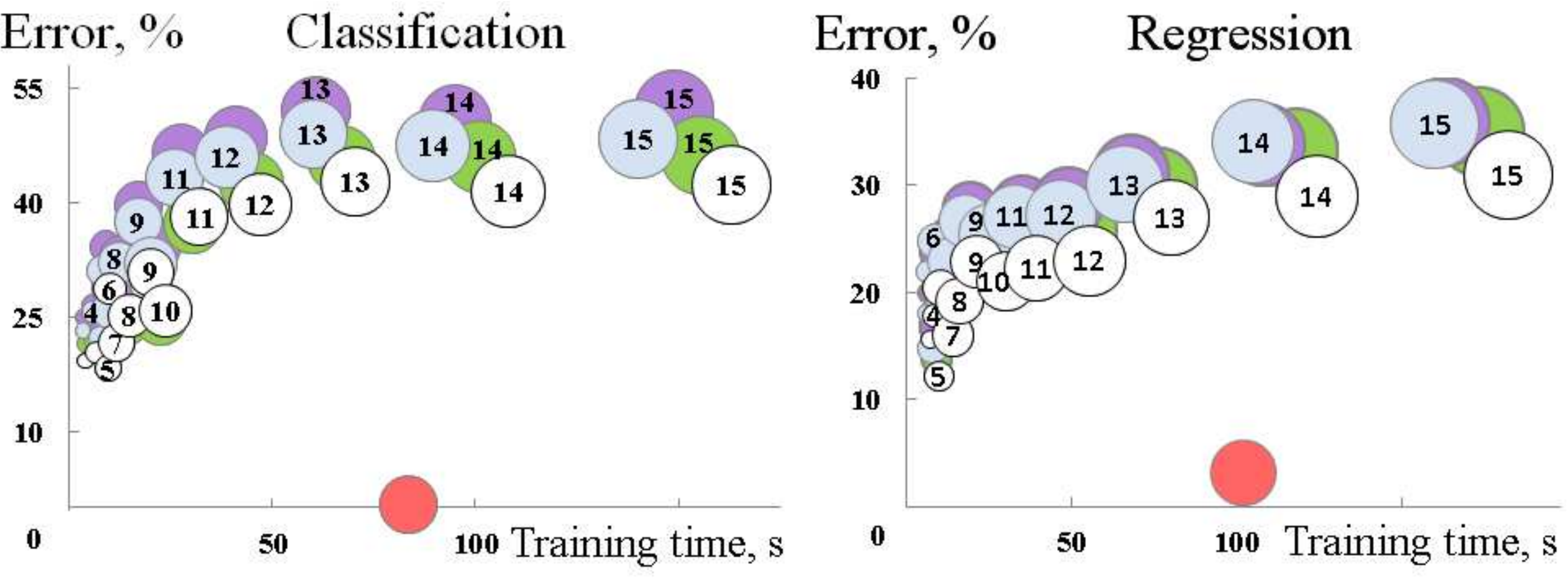}
	\definecolor{GColor}{RGB}{212,226,244}
	\definecolor{VGSColor}{RGB}{146,208,80}
	\definecolor{IColor}{RGB}{168,110,212}			
	\caption[ ]{Prediction error and training time for Decision Stream algorithm (\tikzcircle[gray,fill=DSColor]{4.4pt}) and decision trees with depths ranging from 3 to 15 (are denoted on the labels). X-axis~--- training time, Y-axis~--- prediction error for classification (left) and regression (right) tasks. Decision tree metrics: \tikzcircle[gray,fill=IColor]{4.2pt}~--- information gain, \tikzcircle[gray,fill=GColor]{4.2pt}~--- Gini impurity,  \tikzcircle[black,fill=white]{4.2pt}~--- variance reduction, \tikzcircle[gray,fill=VGSColor]{4.2pt}~--- variance reduction and Gini impurity for continuous and categorical features, respectively.}\label{fig:error:distributed}
\end{figure}
Fig.~\ref{fig:error:distributed} shows the classification error, the regression weighted absolute percentage error (Eq.~\ref{equ:mean:percent}) and the training time for DT with a depth ranging from 3 to 15 levels, and DS which depth is regulated automatically. According to the results, decision trees trained with variance reduction metric and depth restriction of 5 levels demonstrate the best accuracy in both classification and regression tasks and so are used in our further experiments. 
\begin{figure*}
	\centering
	\includegraphics[height=5.9cm]{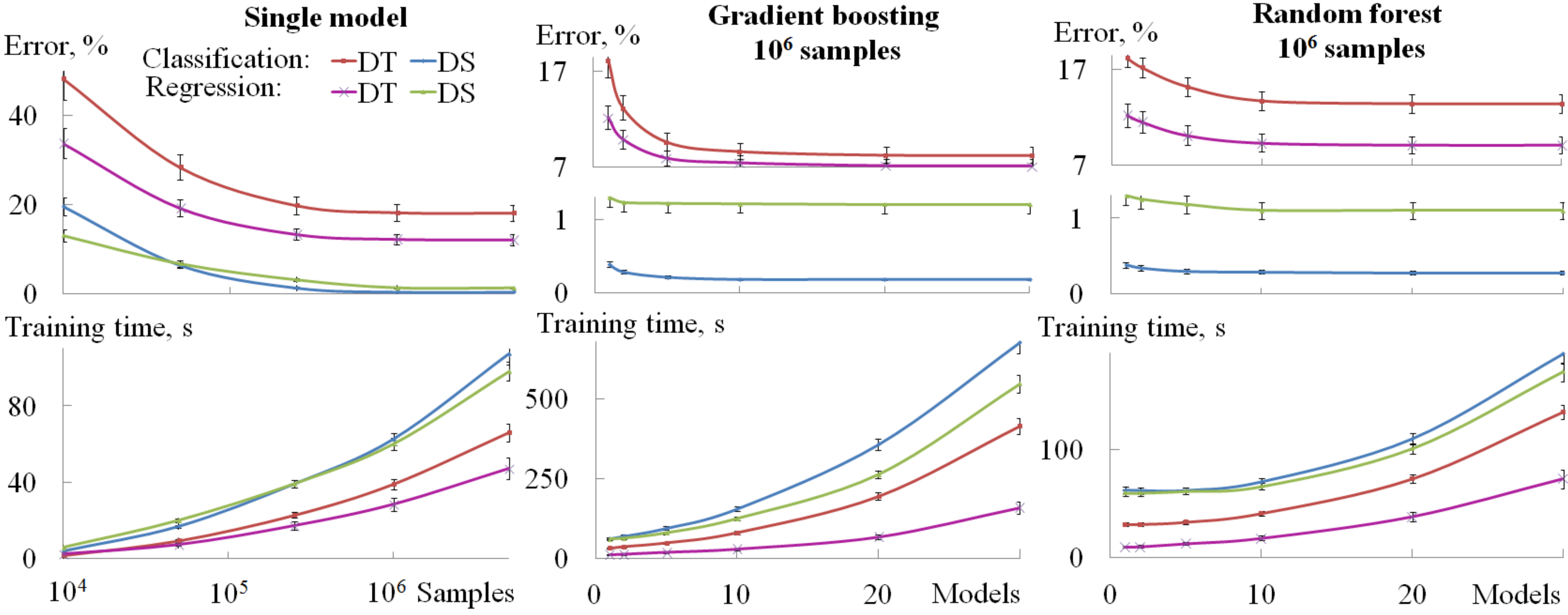}
	\caption{The training time and prediction error ($\pm$\,95\,\% confidence interval) on large-scale data for decision tree (DT), Decision Stream (DS) and their ensembles generated with gradient boosting and random forest algorithms.}\label{fig:error2:distributed}
\end{figure*}

The prediction error of Decision Stream algorithm (\tikzcircle[gray,fill=DSColor]{4.2pt}) is 9~--- 48 times lower than the error obtained by DT. The explanation of this significant difference is in the fact that the generated synthetic data had a distribution that was close to normal, thus the used pair of Z-test/t-test was especially effective in this case. Another reason is that better accuracy was also obtained at the expense of higher running time of DS algorithm.

To find the time that is required for DS and DT to provide the same accuracy, and to compare the accuracy after corresponding training periods, the experiments with different quantity of training data and number of models in ensembles were carried out (Fig.~\ref{fig:error2:distributed}). According to the empirical results presented in Table~\ref{table:error:distributed}, it takes significantly lower amount of data and less training time for DS to provide the same quality of prediction as for DT in both classification and regression tasks; for comparable training time Decision Stream demonstrates significantly better accuracy.

Gradient boosting and random forest ensembles improve DT performance, though the minimal error of ensembles with 30 decision trees is still higher than the corresponding error of 30 Decision Streams: the difference reaches 46~--- 48 times for classification and 5.9~--- 8.3 times for regression tasks. Thus, the proposed modification of Decision Stream for large-scale data demonstrates faster training and better accuracy on both regression and classification tasks compared to DT algorithm.

\begin{table}[h]
	\caption{The training time and prediction error ($\pm$\,95\,\% confidence interval) for decision tree (DT) and Decision Stream (DS) trained on large-scale data.}\label{table:error:distributed}
	\centering
	\newcolumntype{C}{ >{\centering\arraybackslash} m{2.9cm} }
	\newcolumntype{S}{ >{\centering\arraybackslash} m{2.9cm} }	
	\newcolumntype{N}{ >{\centering\arraybackslash} m{2.95cm} }
	\newcolumntype{A}{ >{\centering\arraybackslash} m{0.4cm} }
	\begin{tabular}{A m{1.4cm} S N C C C C}\arrayrulecolor{gray}
		\hline
		& Condition & Model & Samples
		& Time, s & Error, \%\\
		\specialrule{.1em}{0em}{0em}
		
		\multirow{9}{4em}{\rotatebox[origin=c]{90}{Classification}} & \multirow{2}{6em}{Same number of samples} & DT & $10^6$ & 39.2\,$\pm$\,2.31 & 18.2\,$\pm$\,2.33\\[4pt]
		& & DS & $10^6$ & 62.4\,$\pm$\,3.14 & \textbf{0.38\,$\pm$\,0.08}\\
		&\GC &\GC Ratio &\GC 1 &\GC 0.63 &\GC \textbf{48}\\
		
		& \multirow{2}{5em}{Similar time} & DT & $25\cdot10^4$ & 22.7\,$\pm$\,1.62 &19.9\,$\pm$\,2.43 \\
		& & DS & \textbf{5}\,$\cdot$\,\textbf{10}$^4$ & \textbf{16.7\,$\pm$\,1.12} & \textbf{6.36\,$\pm$\,0.83} \\
		&\GC &\GC Ratio &\GC \textbf{5} &\GC \textbf{1.36} & \GC \textbf{3} \\
		
		& \multirow{2}{6em}{Same accuracy} & DT & $25\cdot10^4$ & 22.7\,$\pm$\,1.62 & 19.8\,$\pm$\,2.40 \\
		& & DS & \textbf{10}$^4$ & \textbf{3.82\,$\pm$\,0.23} & 19.6\,$\pm$\,2.28\\	
		&\GC &\GC Ratio &\GC \textbf{25} &\GC \textbf{6} &\GC 1\\
		
		\hline
		
		\multirow{9}{4em}{\rotatebox[origin=c]{90}{Regression}} & \multirow{2}{6em}{Same number of samples} & DT & $10^6$ & 28.3\,$\pm$\,3.43 & 12.2\,$\pm$\,1.28 \\[4pt]
		& & DS & $10^6$ & 60.1\,$\pm$\,2.99  & \textbf{1.32\,$\pm$\,0.12}\\
		&\GC &\GC Ratio &\GC 1 &\GC 0.47 &\GC \textbf{9}\\	
		
		& \multirow{2}{5em}{Similar time} & DT & $10^6$ & 28.3\,$\pm$\,3.43 & 12.2\,$\pm$\,1.28 \\
		& & DS &\textbf{5}\,$\cdot$\,\textbf{10}$^4$ & \textbf{19.9\,$\pm$\,1.29} & \textbf{6,68\,$\pm$\,0.72} \\
		&\GC &\GC Ratio &\GC \textbf{20}  &\GC \textbf{1.42} &\GC \textbf{1.83} \\			
		
		& \multirow{2}{6em}{Same accuracy} & DT & $25\cdot10^4$& 17.1\,$\pm$\,2.15 & 13.2\,$\pm$\,0.13 \\
		& & DS &\textbf{10}$^4$  &\textbf{5.6\,$\pm$\,0.31} & 13.1\,$\pm$\,0.11\\	
		&\GC &\GC Ratio &\GC \textbf{25} &\GC \textbf{3} &\GC 1 \\
		
		\hline
	\end{tabular}
\end{table}

\section{Conclusion}\label{conclusion}

In this paper we presented a novel decision tree based algorithm~--- a Decision Stream, which avoids the problems of data exhaustion and formation of unrepresentative data samples in decision tree nodes by merging the leaves from the same and/or different levels of the predictive model structure. By increasing the number of samples in each node and reducing the tree width, the proposed algorithm preserves statistically representative data and allows extremely deep graph architecture that can consist of hundreds of levels. The main parameter of the algorithm~--- significance threshold, determines the results of each split/merge operation and automatically defines the depth of the Decision Stream model.

The experiments demonstrated that Decision Stream algorithm shows a strong advantage over the standard decision tree learning methods on both regression and classification tasks in both versions: non-distributed for relatively small datasets, where a precise selection of the best data splits is crucial; and distributed, where a balance between the accuracy and computational performance should be maintained.

\end{document}